\pdfoutput=1

\documentclass[11pt]{article}

\usepackage[final]{EMNLP2023}

\usepackage{times}
\usepackage{latexsym}
\usepackage{float}

\usepackage{booktabs, multirow} 
\usepackage{soul}

\usepackage[T1]{fontenc}

\usepackage[utf8]{inputenc}

\usepackage{microtype}

\usepackage{inconsolata}
\usepackage{tcolorbox}

\usepackage{todonotes}

\usepackage{pifont}
\definecolor{bettergreen}{RGB}{52, 168, 63}
\definecolor{betterred}{RGB}{234, 68, 63}
\definecolor{betteryellow}{RGB}{240, 200, 63}
\newcommand{\cmark}{\textcolor{bettergreen}{\ding{51}}}%
\newcommand{\omark}{\textcolor{betteryellow}{\ding{51}}}%
\newcommand{\xmark}{\textcolor{betterred}{\ding{55}}}%

\title{Are BabyLMs Second Language Learners?}

\author{Lukas Edman$^{1,2}$\qquad Lisa Bylinina$^{3}$ \qquad Faeze Ghorbanpour$^{1,2}$ \qquad Alexander Fraser$^{2,4}$ \vspace{.2cm}\\ 
$^{1}$Center for Information and Language Processing, LMU Munich \\ 
$^{2}$Munich Center for Machine Learning \\
$^{3}$Institute for Language Sciences, Utrecht University \\
$^{4}$School of Computation, Information and Technology, TU Munich \\
\vspace{.1cm} {\tt \small lukas@cis.lmu.de, e.g.bylinina@uu.nl, faeze.ghorbanpour@lmu.de, alexander.fraser@tum.de}
}

\begin{document}
\maketitle
\begin{abstract}


This paper describes a linguistically-motivated approach to the 2024 edition of the BabyLM Challenge \cite{warstadt-et-al-2023-babylm}. Rather than pursuing a first language learning (L1) paradigm, we approach the challenge from a second language (L2) learning perspective. In L2 learning, there is a stronger focus on learning explicit linguistic information, such as grammatical notions,  definitions of words or different ways of expressing a meaning. This makes L2 learning potentially more efficient and concise. 
We approximate this using data from Wiktionary, grammar examples either generated by an LLM or sourced from grammar books, and paraphrase data.
We find that explicit information about word meaning (in our case, Wiktionary) does not boost model performance, while grammatical information can give a small improvement. The most impactful data ingredient is sentence paraphrases, with our two best models being trained on 1) a mix of paraphrase data and data from the BabyLM pretraining dataset, and 2) exclusively paraphrase data. 

\end{abstract}

\section{Introduction}

Language models (LMs) need a lot of data in order to learn to approximate human linguistic behaviour \citep{warstadt2022artificial}. The amounts of linguistic data typically used for training recent LMs is is significantly larger than what is available for most of languages of the world, and also much more than what children are typically exposed to during their first language acquisition. A 13 year old is typically exposed to less than 100 million words of linguistic input, which is orders of magnitude less than the amount used in LM pretraining. And still, LMs fail to be quite as good in language as human learners. Can we teach our models to be more data-efficient? If yes, how? 

There are two potential strategies. One is to study how children acquire language in a natural setting, and use their acquisitional trajectories and patterns as inspiration for LM training.
This intuition is one of the motivations for the BabyLM Challenge (hence the name; other low-resource pre-training contexts are, of course, also relevant): the challenge encourages LM pretraining optimization advancements inspired by human linguistic development
\cite{warstadt-et-al-2023-babylm}.

Another direction is to embrace the obvious differences between LM pretraining and the ways human learners acquire their native language. The architectures of current LMs are dramatically different from human brain anatomy, and training objectives and strategies have only limited psycholinguistic developmental parallels. Finally -- and most importantly for our contribution -- input for first language acquisition by human learners and for LM pretraining is hardly comparable not only when it comes to dataset size. While the amount of strictly linguistic input that children get is small compared to typical LM training data, children get this input in communicative context that LMs lack at the pretraining stage, and it is typically paired with cross-modal data, which is not part of the \texttt{strict-small} track we choose for the BabyLM Challenge. 

At a very high level, taking this second direction means that we look beyond human linguistic and cognitive development for optimization strategies -- or at least, we do not need to expect that those will be the ones that necessarily work best. 

We sharpen this point and contrast language learning in an acquisitionally realistic setting (first-language, or L1, acquisition) -- and language learning in a more artificial setting -- learning a second language, L2; a human activity that also leads to (different levels of) linguistic proficiency but contrasts dramatically with L1 acquisition by children. Almost everything is different: the set-up, the data, typical tasks the learner faces, and very often modality and their combinations.

We choose this particular direction mainly because in the current, second, edition, of the BabyLM Challenge participants are allowed to construct their own datasets within the track word budget. A lot of submissions last year, including ours, experimented with curriculum learning -- different ways to order the same data (see our submission \citet{edman-bylinina-2023-much} as well as the BabyLM 2023 findings \cite{warstadt-et-al-2023-babylm}). These attempts gave only limited results.

This year we instead focus on the effect of choosing different data on LM pretraining. In particular, roughly in line with how people learn foreign languages through explicit linguistic instruction, we divide training data into blocks roughly corresponding to types of linguistic information commonly found in English-as-a-foreign-language courses. We participate in the \texttt{strict-small} track allowing for only 10M words and experiment with four different types of linguistic information:
\begin{itemize}
    \item {\bf Lexical information} (information about word meaning and use), parallel to word learning in L2 acquisition. We use Wiktionary data as a source of this knowledge.
    \item {\bf Grammatical information}, parallel to grammar learning for L2. We try two ways of constructing grammar data: a set of sentences marked with grammar phenomena, and texts of grammar books for L2 English learners. 
    \item {\bf Paraphrasing} has perhaps fewer obvious parallels in L2 learning practice, but is related to the explicit focus on sentential semantics (`different ways to say the same thing') and how different modifications in syntax and vocabulary can preserve and alter the meaning of a sentence, which is a common focus in L2 class discussions and exercises. For this data, we use one of the two SynSCE corpora from \citet{zhang2023contrastive}.
    \item A mix of {\bf unconstrained textual data} that corresponds to various input during language acquisition of any kind, be it L1 or L2 acquisition. For this, we use portions of the BabyLM data provided by the challenge organizers.
\end{itemize}

We find that data on paraphrasing brings in the most significant improvements. Grammatical information is only marginally useful, even though it does come with some improvement, depending on the training set-up. Finally, lexical information does not seem useful for LM pretraining. One cannot be sure what to attribute these results to: the usefulness or lack thereof of particular types of data; the quality of the actual various datasets that we use; or the properties of evaluation used to judge whether a particular type of data is useful.
One way or another, our answer to the question of whether BabyLMs are L2 learners is `only when it comes to certain types of data'.

\section{Data}

\subsection{BabyLM data} 

We make use of data provided by BabyLM organizers for our experiments. 
One of our two submitted models (Contr.) doesn't use BabyLM data at all, while the other one (Half/Half) uses a subset of BabyLM data. In the Half/Half model, we use the following parts of the BabyLM dataset:

\begin{table}[!htp]\centering
\begin{tabular}{lrr}\toprule
Dataset &Words \\\midrule
Simple Wikipedia & 145K \\
Gutenberg & 254K \\
Switchboard &147K \\
\bottomrule
\end{tabular}
\caption{BabyLM data used for the Half/Half model.}\label{tab:data}
\end{table}

We think BabyLM data roughly corresponds to unconstrained linguistic input in a language learner's experience (reading materials and practice conversations with language teachers and peers).

The rest of the data in the Half/Half model comes from the dataset we discuss next. 

\subsection{Contrastive dataset} 

An important part of the language acquisition experience is finding out how changes in phrasing and syntactic structure can alter or preserve meaning. This is seen in typical L2 learning tasks such as paraphrasing, which highlight the semantics of the sentence and the ways syntactic manipulation can affect its meaning. 

As data approximating this type of information, we use a dataset by \citet{zhang2023contrastive}. They release two datasets as part of SynCSE, a contrastive learning framework for training sentence embeddings. The data in both datasets (SynCSE-partial and SynCSE-scratch) is synthetic: synthesized by LLMs. The two different datasets are results of different prompting set-ups (for the dataset construction and prompting details, we refer the reader to the original paper). We use one of these two datasets, SynSCE-partial\footnote{\href{https://huggingface.co/datasets/hkust-nlp/SynCSE-partial-NLI}{https://huggingface.co/datasets/hkust-nlp/SynCSE-partial-NLI}}. 

The dataset is structured as follows: each datapoint comes as a triple consisting of 1) a sentence; 2) its paraphrase, and 3) a hard negative (a sentence that is similar to the original one lexically and/or structurally but has a different meaning). Here is an example of a triplet from the dataset: 

\begin{quote}
    \textbf{sent0:} \texttt{One of our number will carry out your instructions minutely.}
    
\textbf{sent1:} \texttt{One person from our group will execute your instructions with great attention to detail.}

\textbf{hard\_neg:} \texttt{Each member of our group will carry out your instructions differently.}
\end{quote}

We use all three elements of the triplet in our experiments. 

\subsection{Grammar data}
To mimic explicit grammar instruction in the typical L2 learning setting, we look for ways to expose the model to targeted grammatical information. We explore two strategies and corresponding datasets, which we call \texttt{Gram Gen} and \texttt{Gram Books}.

For \textbf{Gram Gen}\footnote{We release this dataset on HF: \texttt{link placeholder}.}, we first compile a list of grammatical notions that a sentence can contain. This list is inspired by the typical structure of reference and learners' grammars and the topics covered by those. We then pass these notions to GPT 4o-mini\footnote{We changed from GPT 3.5 to 4o-mini due to pricing changes.} to generate examples, using the prompt in Figure \ref{fig:gram_prompt}. To ensure that we generate a diverse set of sentences, we prompted the model to generate sentences about specific topics.\footnote{The possible topics are: accounting, anthropology, archaeology, architecture, art, artificial intelligence, astronomy, biology, botany, business, chemistry, computer science, cosmology, criminology, design, economics, education, environmental science, engineering, geography, geology, government, history, humanities, international relations, journalism, law, literature, linguistics, math, medicine, music, philosophy, physics, poetry, politics, psychology, religion, sports, and theater.} 

After this, we additionally tag each sentence with the grammatical notions as a sentence can contain more than one. This again is done with GPT 4o-mini, using the prompt in Figure \ref{fig:gram_tag_prompt}. Due to pricing restrictions, we generate 500 sentences per notion, and tag 100 of these sentences for 50 different notions. We include an example of a sentence tagged, where we verify the correctness of the given tags in Table \ref{tab:tags_ex}. 

In the table we can see that GPT 4o-mini appears only partially capable of recognizing grammatical notions. For the simpler, very well-known notions such as common nouns, verb person, tense, and number, GPT performs well. For less commonly-known phenomena, such as ellipsis, it seems to have no understanding. For ellipsis specifically, GPT often has false positives with sentences of this 2-clause structure, likely because that is a necessary component for an ellipsis to occur, but not what defines an ellipsis. GPT also appears to occasionally hallucinate, with ``it'' not appearing in the sentence despite it being tagged as an object pronoun. Overall, given the accuracy of GPT in tagging, it is not surprising that our model would struggle to grasp grammatical notions. 

\begin{table}[!htp]\centering

\scriptsize
\begin{tabular}{p{2.4cm}p{3.1cm}c}\toprule
\\
\multicolumn{3}{c}{``The engineers proposed a new design for the bridge, while the architects} \\
\multicolumn{3}{c}{ focused on the aesthetic elements, emphasizing sustainability instead.''} \\ 
\\ \toprule
Notion &Tag & Correct? \\\midrule
common noun &engineers, design, bridge, architects, elements, sustainability   & \cmark \\
collective noun &engineers, architects & \omark \\
singular noun &design & \cmark \\
plural noun &engineers, architects, elements & \cmark \\
nominative case &The engineers & \cmark \\
simple past tense &proposed, focused, emphasized & \omark \\
third person &engineers, architects & \cmark \\
plural verb &proposed, focused, emphasizing & \omark \\
indicative mood &proposed, focused, emphasizing & \cmark \\
non-gradable adjective &sustainable  & \omark \\
positive adjective &sustainable & \xmark \\
aspectual adverb &emphasizing & \xmark \\
comparative adverb &instead & \xmark \\
object pronoun &it & \xmark \\
case preposition &for, on, instead & \omark \\
coordinating &while & \cmark \\
indefinite determiner &a new design & \omark \\
noun phrase &The engineers, a new design, the bridge, the architects, the aesthetic elements, sustainability  & \omark \\
adjectival modification &aesthetic, sustainability & \omark \\
verb phrase &proposed, focused, emphasizing & \xmark \\
transitive verb phrase &proposed a new design, focused on the aesthetic elements, emphasizing sustainability & \omark \\
 \\
direct object &design, elements & \cmark \\
adjunct clause &Yes & \cmark \\
ellipsis gapping &Yes & \xmark \\
ellipsis pseudo-gapping &Yes & \xmark \\
\bottomrule
\end{tabular}
\caption{Tags produced for the sentence above. Only positive tags are shown for brevity. \cmark~indicates the tag is completely correct, \omark~partially correct, \xmark~incorrect. }\label{tab:tags_ex}
\end{table}


We construct the second grammar dataset, \textbf{Gram Books}, as an alternative to grammatical instruction via examples. This dataset contains grammar books that overtly discuss the rules of English grammar and are intended mainly for second language learners of English. Here is the full list of the grammar books we used:  \citet{newson2006basic, greenbaum2009introduction, roth2010compelling, thomson2015practical, brutjan2022learn, wright2024english}. We do not release this dataset due to copyright constraints.

\begin{figure}
    \centering
\begin{tcolorbox}
You are an expert in grammar. Write 500 detailed sentences containing <notion> (as opposed to <alternate notion>). Make sure to write 500 detailed sentences that are all different from each other. Try to make the sentences sufficiently different, for example, don't start every sentence with ``the'', make both short and long sentences, and write about the topic of <topic>. Don't write anything else.
\end{tcolorbox}
    \caption{The prompt used to generate example sentences of a grammatical notion. The <alternate notion> is not always used, but corresponds to notions with clear alternatives, such as telic vs. atelic verbs.}
    \label{fig:gram_prompt}
\end{figure}

\begin{figure}
    \centering
\begin{tcolorbox}
Consider the sentence: <sentence> 
Does the sentence contain the notion of <notion>? If so, write which word or words correspond to the notion. If not, write ``N/A''. Only write the word or words that correspond, or N/A otherwise.
\end{tcolorbox}
    \caption{The prompt used to tag sentences with their grammatical notion. The prompt for sentential notions only contained the initial question, along with: ``Answer with yes or no. Only write `yes' or `no', nothing else.''}
    \label{fig:gram_tag_prompt}
\end{figure}

We use both grammar datasets for two types of experiments: 1) regular MLM training (described in Section \ref{training}); 2) more elaborate training schemes involving a combination of an encoder and a decoder (discussed in Section \ref{other}).

\subsection{Wiktionary} 

For lexical instruction, we make use of a segment of data from Wiktionary\footnote{\href{http://www.wiktionary.org/}{http://www.wiktionary.org/}}, the largest available collaborative source of lexical knowledge. We constrain ourselves to the English segment of Wiktionary, and extract the lemma together with parts of speech and the definitions of each of its senses and examples that illustrate the senses. 

We parse the Wiktionary data into CSV, where each row contains a word, part of speech, a definition, and up to 13 examples, though many contained no examples. 

For words without an example, we attempted two things: we generated examples with GPT 3.5, and we fed the word in as is. The examples generated were of notably high quality, with GPT even able to generate sentences for rare word senses. The prompt we used is shown in \ref{fig:wikt_prompt}.

\begin{figure}
    \centering
\begin{tcolorbox}
Give 3 examples of the word <word> as a(n) <part of speech>, where it means <definition>. List the 3 examples in a numbered list, they should be full sentences. Don't say anything else. The format should look like: \\
1. Example 1 \\
2. Example 2 \\
3. Example 3
\end{tcolorbox}
    \caption{The prompt used to generate example sentences of a word sense. }
    \label{fig:wikt_prompt}
\end{figure}

As with other types of linguistic knowledge, with this data we are looking for a way to mimic typical L2 learning. Wiktionary comes pretty close to word learning in this setting, as it contains explicit information about different senses of the word, its morphological and syntactic profile, defines its lexical semantics and illustrates all of this information with sentences where the word is used in its different senses.

Again, as with grammar data, we use the resulting Wiktionary dataset\footnote{The dataset we construct is available on HF: \texttt{link placeholder}.} both in experiments with simple MLM pretraining and in experiments with more complicated training set-ups, which are described in more detail in Sections \ref{training} and \ref{other}, respectively.

\section{Method}
\subsection{Model Choice}
We opted to use encoder-only models for our final submission. This is based on our observation from last year's competition, where encoder-only models generally outperformed decoder-only or encoder-decoder models. We chose the DeBERTa-base \cite{he2021debertav3} architecture as it is considered state-of-the-art for encoder-only models. Unlike in last year's competition where we saw improvements from using DeBERTa-large, we saw no improvement this year in initial testing and thus only used the base model size. 

\subsection{Training and Evaluation}\label{training}
Our pretraining uses the standard MLM scheme \cite{liu2019roberta}, which we used last year to great effect. Table \ref{tab:hyperparams} shows the hyperparameters we used for our pretraining experiments. For fine-tuning, we use the default hyperparameters provided by the organizers.

\begin{table}[!htp]\centering
\begin{tabular}{lrr}\toprule
Hyperparameter &Value \\\midrule
Vocabulary size & 40000 \\
Context size & 64 \\
Learning rate &2e-4 \\
Decay &0.01 \\
Warmup steps & 4000 \\
Optimizer &AdamW \\
Batch size & 64, 256 \\
Epochs &50 \\
\bottomrule
\end{tabular}
\caption{Hyperparameters used.}\label{tab:hyperparams}
\end{table}

The hyperparameters chosen are largely the same as what we used in last year's competition \cite{edman-bylinina-2023-much}, with some minor changes to the learning rate (2e-4 vs. 1e-4) and warmup steps (4000 vs. 10000), as well as using both a batch size of 64 and 256. We found that, in some circumstances, a batch size of 64 would result in a more performant model, but this phenomenon was inconsistent. As such, we report the best performing batch size for each model. We note that ``context size'' refers to the number of tokens in a given example. This is constant, so each example may contain multiple sentences or fragments.

We evaluate our models with the tasks included in this year's shared task: BLiMP \citep{warstadt2020blimp}, BLiMP supplement, (Super-)GLUE \citep{wang2018glue,wang2019superglue}, and EWoK \citep{ivanova2024elements}.

\subsection{Additional Training Schemes}\label{other}
In addition to using encoder-only MLM training, we experimented with other objectives to train using our Wiktionary and grammar data, but ultimately found no discernible difference in performance. For these experiments, we use an encoder-decoder model, where the decoder is later removed after training. The encoder part is simultaneously trained on MLM as well as the additional objectives, which we now describe.

\begin{figure}
    \centering
    \includegraphics[width=\linewidth]{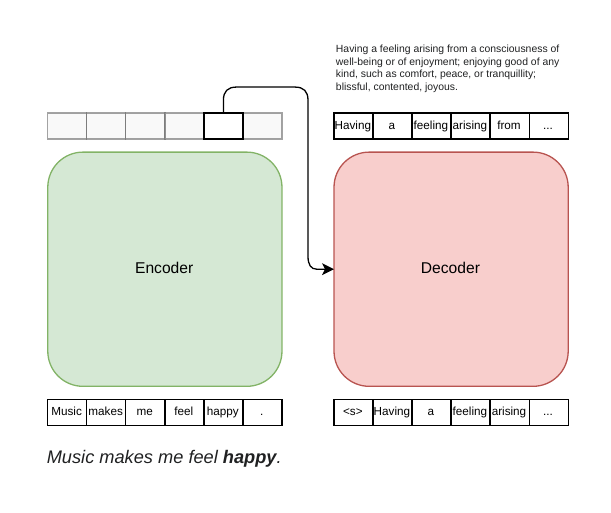}
    \caption{The model layout for training wiktionary.}
    \label{fig:wikt_model}
\end{figure}

\paragraph{Wiktionary Training}
For each Wiktionary entry, we feed the example as input to the encoder and mark the specific token that corresponded to the target word. For the marked position, we pass this to a separate decoder, which is tasked with generating the definition. This process can be seen in Figure \ref{fig:wikt_model}.

\paragraph{Grammar Training} \label{grammar}

\begin{figure}
    \centering
    \includegraphics[width=\linewidth]{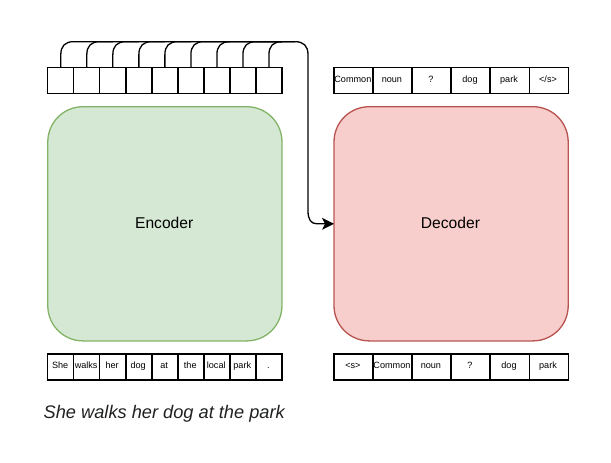}
    \caption{The model layout for training with grammar examples.}
    \label{fig:gram_model}
\end{figure}

For the \texttt{Gram Gen} data, we feed in a sentence to the encoder, passing its hidden states to the decoder, and prompt the model to answer whether it contains a particular notion, and if that notion corresponds to a particular word or words, which word(s) does it correspond to. The scheme for training is shown in Figure \ref{fig:gram_model}.

\section{Results}
We first discuss the results of our experiments with MLM-only models trained on grammar and lexical data, then we move on to discuss the results of the models with additional training schemes. Finally, we cover the results of our best-performing models that we submitted to the challenge.

\subsection{Grammar Results}

The results for our best models using grammar data are shown in Table \ref{tab:grammar}. As we can see, adding grammar data appears to help with BLiMP to a limited extent, but hurts performance on all other metrics. The increase in BLiMP is expected, as the BLiMP evaluation necessitates that grammatical sentences are given a lower perplexity than ungrammatical sentences. A lot of the sentences in BLiMP 
are 
grammatical, but are very unnatural for a native speaker to read. As such, an excellent source for unnatural sounding yet grammatically correct sentences is a grammar book. This is likely why we see the most improvement from training on those. 

The generated data, seeing as it is generated by GPT 3.5, is likely going to reflect the data that GPT itself was trained on. Although we do not know specifically the data that GPT is trained on, it is likely much more representative of ``natural'' data, rather than these unnaturally constructed sentences that are ubiquitous in BLiMP.


\begin{table}[!htp]\centering

\scriptsize
\begin{tabular}{lrrrr}\toprule
&Half / Half &+ Gram Gen &+ Gram Books \\\midrule
BLiMP &74.2 &74.7 &\textbf{75.4} \\
Supplement &\textbf{63.7} &63.3 &61.1 \\
GLUE &\textbf{77.1} &75.9 & 74.7 \\
EWoK &\textbf{70.0} &67.7 &66.1 \\ \midrule
Average &\textbf{71.3} &70.4 & 69.3 \\
\bottomrule
\end{tabular}
\caption{Results of our grammar-informed models.}\label{tab:grammar}
\end{table}

To further improve BLiMP scores, we expect that including more grammar books or perhaps explicitly prompting an LLM to produce unnatural sounding sentences may be the key. However, we also expect that such data would have a negative impact on GLUE and EWoK. This may simply be an immutable trade-off for low-resource pretrained models.

\subsection{Wiktionary Results}

We show the results of adding Wiktionary data in Table \ref{tab:wikt}. Unfortunately, adding Wiktionary definitions and examples appears to only hurt performance. We speculate that it might have to do with the structure of Wiktionary entries and how the structure of lexical information is drastically different from other types of training and evaluation data.

\begin{table}[!htp]\centering

\scriptsize
\begin{tabular}{lrrr}\toprule
&Half / Half &+ Wikt \\\midrule
BLiMP &\textbf{74.2} &72.9 \\
Supplement &\textbf{63.7} &62.8 \\
GLUE &\textbf{77.1} & 75.7 \\
EWoK &\textbf{70.0} &65.8 \\ \midrule
Average &\textbf{71.3} &69.3 \\
\bottomrule
\end{tabular}
\caption{Results of adding Wiktionary data.}\label{tab:wikt}
\end{table}

\subsection{Additional Training Schemes Results}

\begin{table}[!htp]\centering
\scriptsize
\begin{tabular}{lrrrr}\toprule
&MLM &MLM + Gram &MLM + Wikt \\\midrule
BLiMP &74.2 &71.5 &\textbf{75.7} \\
Supplement &\textbf{63.7} &61.0 &59.3 \\
GLUE &\textbf{77.1} &75.9 &73.4 \\
EWoK &\textbf{70.0} &67.7 &66.5 \\ \midrule
Average &\textbf{71.3} &69.0 &68.7 \\
\bottomrule
\end{tabular}
\caption{Our models with additional objectives, compared to the MLM-only baseline (i.e. our half/half model).}\label{tab:extra_results}
\end{table}

We show the results of our models with added objectives for Wiktionary definition learning and grammatical notion identification in Table \ref{tab:extra_results}.
Concerning the grammar objective, we see slightly worse performance overall. Notably, despite BLiMP being an evaluation aimed at gauging understanding of grammaticality, we still see a decrease in the performance. 

\begin{table*}[!htp]\centering
\scriptsize
\begin{tabular}{lrrrrrrr}\toprule
&\multicolumn{2}{c}{BabyLlama} &\multicolumn{2}{c}{LTG-BERT} &BabyLM &Half / Half &Contr. \\\cmidrule(lr){2-3} \cmidrule(lr){4-5} \cmidrule(lr){6-6} \cmidrule(lr){7-7} \cmidrule(lr){8-8}
&10M &100M &10M &100M & 10M & 10M & 10M \\\midrule
BLiMP &69.8 &73.1 &60.6 &69.2 &\textbf{74.2} &\textbf{74.2} &65.5 \\
Supplement &59.5 &60.6 &60.8 &\textbf{66.5} &66.2 &63.7 &60.3 \\
GLUE &50.7 &52.1 &48.9 &51.9 &69.0 &\textbf{77.1} &76.6 \\
EWoK &63.3 &69.0 &60.3 &68.4 &67.5 &\textbf{70.0} &67.3 \\ \midrule
Average &60.8 &63.7 &57.7 &64.0 &69.2 &\textbf{71.3} &67.4 \\
\bottomrule
\end{tabular}
\caption{Final results compared to the baselines.}\label{tab:final}
\end{table*}

Ironically, our Wiktionary-based objective increases BLiMP scores. It is unclear why our method for improving semantic understanding increased performance on the grammar benchmark, but there is of course information that can be extracted from word definitions that is useful for parsing grammaticality, such as part of speech information, and even quite literal information about the usage of words (e.g. the definition of ``the'' starts with ``used before a noun phrase...").

Though it does not explain the improvement on BLiMP from our model trained with the Wiktionary objective, we believe that adding an additional objective is the main source of the loss in performance for our additional models. BLiMP (as well as EWoK) is designed such that a model's zero-shot default behavior is to provide a perplexity for a sentence. This is achieved trivially with a model trained on MLM or CLM, but adding another objective means that the hidden states are forced to learn a representation that balances approximating the perplexity with optimizing for whatever the external objective requires. Thus, it is no surprise that the scores for BLiMP and EWoK are lower. This does not necessarily mean that this model is less capable of understanding grammaticality, but this could not be captured by BLiMP. We are not aware of another benchmark that would resolve this issue.

\subsection{Submission}

In Table \ref{tab:final}, we show the overall results for our best models, compared to the baselines. The results from BabyLlama and LTG-BERT are taken from the reported scores from the organizers. The ``BabyLM'' model is our internal baseline, using the same parameters and training as our other models, but trained on the data provided by the organizers. ``Half / Half '' is a model trained on a mixture of the provided data and contrastive data, and ``Contr.'' is trained on exclusively contrastive data. 

As we can see, our models outperform even the provided models trained on 100M overall.  We suspect this is for the same reason as we found last year in \citet{edman-bylinina-2023-much}, where the models trained on too large of a context size have trouble converging. In terms of the data used, we see that using the contrastive dataset hurts BLiMP performance, but raises GLUE performance. Using a mix is able to capture a best of both worlds, retaining performance on BLiMP while even improving performance on GLUE and EWoK.

\section{Conclusion}

In this year's BabyLM Challenge, we attempted to buck the trend of administering strategies based on L1 acquisition, having seen little success from such strategies in last year's Challenge. Instead, we hypothesized that L2 acquisition, with more explicit information regarding semantics and syntax, might be what a language model needs. To that end, we also saw limited success. Our strategy of using Wiktionary data did not show any indication of improved output quality. Using grammar information did have a small positive effect on BLiMP scores, though it is unclear whether the grammar itself helped or simply the more diverse data domain. 

Nevertheless, our strategy of reducing context size from the previous year was yet again successful at outperforming the baselines, even those with 10$\times$ more data used in training. Additionally, using data that includes paraphrases and contrastive pairs helped improve the GLUE scores by a remarkable 8 points. This goes to show that the data chosen for low-resource pretraining can have a profound impact. The study of the exact structure of data that LMs efficiently learn from is a productive future direction, as tentatively shown by our results.






\section*{Acknowledgements}
We thank BabyLM anonymous reviewers for useful comments. We also thank Oleg Serikov for helpful informal discussions. The work was supported by the European Research Council (ERC) under the European Union's Horizon Europe research and innovation programme (grant agreement No. 101113091) and by the German Research Foundation (DFG; grant FR 2829/7-1).

\bibliography{anthology,custom}
\bibliographystyle{acl_natbib}


\end{document}